\documentclass{article}

\usepackage[accepted]{icml2019}
\usepackage{booktabs}
\usepackage{hyperref}
\usepackage{url}
\usepackage[ruled,vlined]{algorithm2e}
\usepackage[pdftex]{graphicx}
\usepackage{amsmath}
\usepackage[noend]{algpseudocode}
\usepackage{booktabs,tabularx, ragged2e}
\usepackage{multirow}
\newcolumntype{Y}{>{\raggedright\arraybackslash}X} 
\newcolumntype{R}[1]{>{\RaggedLeft\arraybackslash}p{#1}}

\usepackage{url}
\usepackage{subfig}
\usepackage{balance}

\usepackage{wrapfig}

\usepackage[all]{nowidow}
\newcommand{\Fonescore}{$F_1$\xspace}

\icmltitlerunning{Biomedical Named Entity Recognition via Reference-Set Augmented Bootstrapping}

\begin{document}

\twocolumn[
\icmltitle{Biomedical Named Entity Recognition via Reference-Set Augmented Bootstrapping}
%\icmlsetsymbol{equal}{*}

\begin{icmlauthorlist}
\icmlauthor{Joel Mathew}{isi}
\icmlauthor{Shobeir Fakhraei}{isi}
\icmlauthor{Jos\'{e} Luis Ambite}{isi}
\end{icmlauthorlist}

\icmlcorrespondingauthor{Jos\'{e} Luis Ambite}{ambite@isi.edu}

\icmlaffiliation{isi}{Information Sciences Institute, University of Southern California, 4676 Admiralty Way, Marina del Rey, CA}

\icmlkeywords{Biomedical, NER, Natural Language Processing, Machine Learning}

\vskip 0.3in
]

\printAffiliationsAndNotice{}  % leave blank if no need to mention equal contribution
% \printAffiliationsAndNotice{\icmlEqualContribution} % otherwise use the standard text.

\begin{abstract}
We present a weakly-supervised data augmentation approach to improve Named Entity Recognition (NER) in a challenging domain: extracting biomedical entities (e.g., proteins) from the scientific literature. First, we train a neural NER (NNER) model over a small seed of fully-labeled examples. Second, we use a reference set of entity names (e.g., proteins in UniProt) to identify entity mentions with high precision, but low recall, on an % a large
unlabeled corpus. Third, we use the NNER model to assign weak labels to the corpus. Finally, we retrain our NNER model iteratively over the augmented training set, including the seed, the reference-set examples, and the weakly-labeled examples, which improves model performance. We show empirically that this augmented bootstrapping process significantly improves NER performance, and discuss the factors impacting the efficacy of the approach.
\end{abstract}

%%%%%%%%%%%%%%%%%%%%
%%% INTRODUCTION %%%
%%%%%%%%%%%%%%%%%%%%

\section{Introduction}
The increasing wealth of available biomedical data fuels numerous machine learning applications. Unfortunately, much of this data is unlabeled, unstructured and noisy. Supervised learning achieves the best task performance, but obtaining training labels is expensive. Crowd-sourcing could provide labels at scale, but may not be appropriate for acquiring high-quality labels in technical domains that require expert annotators, such as biomedicine. In this paper, we explore augmented bootstrapping methods that leverage automatically assigned noisy labels obtained from a large unlabeled corpus.

The biomedical literature is a high-impact domain with scarce annotations. Unlocking the knowledge in this data requires machine reading systems that automatically extract important concepts in the text, such as entities and their relations. A critical component of such systems is reliable Named Entity Recognition (NER), which aims to identify parts of the text that refer to a named entity (e.g., a protein). In line with advancements in many domains, most state-of-the-art NER approaches use a deep neural network model that relies on a large labeled training set~\citep{fakhraei2018nseen}, which is not usually available in biomedical domains. 
To address label scarcity, we propose a framework to train any effective neural NER model by leveraging partially labeled data. We do this by creating an augmented training set using a small fully-labeled \textit{seed} set, and an unlabeled \textit{corpus} set, which we weakly and automatically label, and then refine its labels via an iterative process.

Our main \textbf{contributions} include:
(1) An augmented bootstrapping approach combining information from a reference set with iterative refinements of soft labels to improve NER performance in a challenging domain (biomedicine) where manual labelling is expensive. 
(2) An analysis of factors affecting performance in a controlled experimental setup.
(3) An analysis of reference-based automated approaches to labeling data, showing that naive labeling decreases performance and how to overcome it.

%%%%%%%%%%%%%%%%%%%%
%%% RELATED WORK %%%
%%%%%%%%%%%%%%%%%%%%

\section{Related Work}
Many effective NER systems assume a fully-supervised setting to train a neural network model \citep{Liu2018EmpowerSL,ma2016end,lample2016neural}. 
Recently, distant supervision has been applied to language-related tasks such as phrase mining \citep{Shang2018AutomatedPM}, relation extraction \citep{mintz2009distant}, and entity extraction \citep{he2017autoentity}. For NER,  \citet{fries2017swellshark} automatically generated candidate annotations on an unlabeled dataset using weak labellers. \citet{ren2015clustype} and \citet{he2017autoentity} used knowledge bases and linguistic features to tag entities. 
Our approach combines knowledge extracted from an external reference set with noisy predicted labels and refines them iteratively. 

Using a reference set, \citet{Ratner2017SnorkelRT} proposed heuristic-based functions to label data with low accuracy. \citet{Shang2018AutomatedPM, shang2018learning} described techniques to automatically tag phrases based on biomedical knowledge bases, such as MeSH and CTD . 

However, in NER systems with weak supervision, wrongly-labeled entities negatively affects the overall performance~\citep{shang2018learning}. 
Our proposed iterative training technique is able to make the learning process more robust to noisy labels.

%Bootstrapping
Our method is closely related to bootstrapping. 
\citet{yarowsky1995unsupervised} introduced the bootstrapping technique by training a tree-based classifier for word-sense disambiguation on labeled seed data and then using it to predict on an unlabeled corpus which further is used for training the model iteratively until convergence. Later \citet{kozareva2006bootstrapping}  bootstrapped statistical classifiers for NER. 
\citet{abney2004understanding} and \citet{Haffari2007AnalysisOS} applied bootstrapping for language processing, and \cite{Reed2015TrainingDN} for image classification. 

We propose an augmented bootstrapping technique for the state-of-the-art neural NER model applied to biomedical literature. In contrast to standard bootstrapping techniques that use hard labels, we refine soft label values, which may be more suitable for noisy data.  
More importantly, we further augment the bootstrapping process via a simple domain-independent data annotation scheme based on a reference set, which is in contrast to the hand-crafted domain specific rules or the linguistic or morphological characteristics used in standard bootstrapping approaches. 

%%%%%%%%%%%%%%
%%% METHOD %%%
%%%%%%%%%%%%%%

\section{Reference-set Labelling and Augmented Bootstrapping} 
\label{method}
Our main goal is to use easily available external information to leverage unlabeled data and reduce the need for an expensive, fully-labeled dataset. 
We assume to have a small fully-annotated \textit{seed} dataset $\mathcal{D}_s$ that has every token tagged by entity type
and a larger \textit{unlabeled corpus} $\mathcal{D}_c$. 
We seek to automatically generate an augmented dataset by partially, and possibly noisily, labeling $\mathcal{D}_c$.
We show that training a (Neural) NER system over the combined seed and augmented datasets achieves the performance of systems trained with an order of magnitude more labels. 

\subsection{Leveraging Reference Sets and Iterative Label Refinement}

We propose an iterative solution to improve NER by labeling the corpus dataset using two complementary sources of information. 
First, we train an NER model using the small seed dataset $\mathcal{D}_{s}$ and use it to label the unlabeled corpus $\mathcal{D}_{c}$; we call this set of labels \textit{predicted labels}. Second, we use \textit{search policies} over a \textit{reference set} to find mentions of entity names in the unlabeled corpus $\mathcal{D}_{c}$; we call these set of labels \textit{reference-based labels}. We combine the seed, the predicted and the reference-based labels to fine-tune the NER model by resuming training from the previous model state. We then use the updated model to iteratively refine the \textit{predicted labels} portion of the corpus set.

\begin{figure}[tbp]
  {\centering
  \includegraphics[width=\columnwidth]{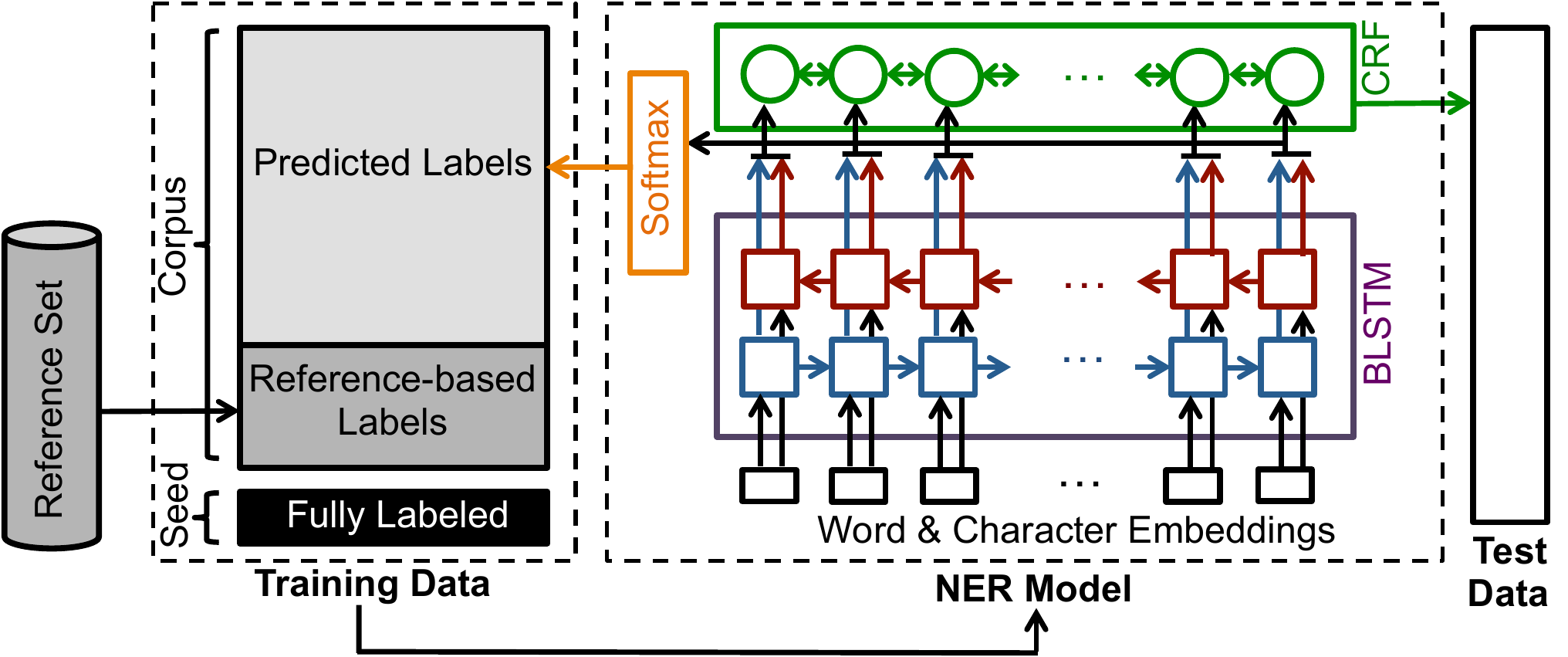}
   \caption{Architecture: NNER+Augmented Bootstrapping.}
   \label{fig:method}}
\end{figure}

Figure~\ref{fig:method} and Algorithm~\ref{alg:method} show the overall process of our method. We use soft scores (between 0 and 1) to label the corpus set, instead of the binary labels produced by the CRF layer used in state-of-the-art NER models.  Our aim is to let the model iteratively reinforce the weak signals in the soft scores to improve the label quality.

      \begin{algorithm}
        \caption{Assignment algorithm}
        \label{alg:method}
        \SetKwProg{Fn}{Function}{}{}
        \SetAlgoNoLine
        \Fn{IterativeTrain ($\mathcal{D}_{s}$, $\mathcal{D}_{c}$)}{
            \KwIn{Labeled seed data ($\mathcal{D}_{s}$)}
            \KwIn{Unlabeled corpus ($\mathcal{D}_{c}$)}
            \KwOut{Iteratively trained model ($\mathcal{M}_{K}$)}
            Train model $\mathcal{M}_0$ on $\mathcal{D}_{s}$
            \\
            \For{i in 1 \ldots K}{
                $\mathcal{D}^{(i-1)*}_{c} \gets$ Predict using $\mathcal{M}_{i-1}$\\
                $\mathcal{D}^{(i-1)}_{c} \gets$ Relabel $\mathcal{D}^{(i-1)*}_{c}$ \\ \textbf{s.t.}\\
                \If{token $\in$ Reference Set}{
                $score_{tag}(token) \gets$ 1
                }
                Train model $\mathcal{M}_{i}$ on $\mathcal{D}_{s}$ + $\mathcal{D}^{(i-1)}_{c}$
                }
            \Return{$\mathcal{M}_{K}$}
            }
      \end{algorithm}

\subsection{Base NER Model and Soft Labeling}

Recent high-performing neural NER (NNER) models \citep{lample2016neural, ma2016end} use Bi-directional LSTM (BiLSTM) layers trained on character and word embeddings. The character embeddings are learned over the training data and concatenated with GloVe word embeddings \citep{pennington2014glove}. The final tag scores are passed to a linear chain CRF layer which produces the most probable sequence of tags.
We use an open-source Tensorflow implementation of this model \citep{guillaumener}, which achieves state-of-the-art NER performance on the CoNLL 2003\footnote{https://www.clips.uantwerpen.be/conll2003/ner/} dataset. 
To produce soft scores for each tag in our experiments, we replace the CRF layer with a \textit{softmax} layer. Entities found via the reference set receive a score of~1. In order to evaluate our initial predictions, we apply \textit{argmax} to the output tag probabilities from the \textit{softmax} layer and generate the most probable labels.
We use the final refined dataset to train an unmodified NNER (BiLSTM+CRF without a Softmax layer) model which has the CRF in the output layer. The code of the model architecture and iterative training will be made available upon paper publication.

%%%%%%%%%%%%%%%%%%%
%%% EXPERIMENTS %%%
%%%%%%%%%%%%%%%%%%%

\section{Experimental Analysis and Results}

We show the effectiveness of our approach in a challenging NER task, extracting protein mentions from the biomedical literature, and systematically evaluate the contribution of the different techniques.

We use the BioCreative VI Bio-ID dataset \cite{arigi2018:bc6bioid}, which contains 13,573 annotated figure captions corresponding to 3,658 figures from 570 full length articles from 22 journals, for a total of 102,717 annotations. The Bio-ID dataset is split into a training set of 38,344 sentences, a development set of 4,243 sentences, and a test set with 14,079 sentences. The tokens are tagged using the \texttt{BIO} scheme (Beginning, Inside and Outside of entities). 

The Bio-ID dataset provides us with a controlled environment where we can evaluate our methods, since it provides ground truth on the labels. The rationale of the following experiments is to simulate our desired data augmentation scenario, which is to search for sentences containing relevant bioentities (e.g., proteins) in a large unlabeled corpus, such as PubMed Central. 
We evaluate our three main techniques, namely (1) using a reference set of entity names (i.e., protein names from UniProt), (2) predicting labels for unknown tokens using a NNER system trained in a small fraction of the data, and (3) refining the label predictions by fine-tuning the NNER system iteratively. We focus on protein/gene annotations for simplicity (51,977 mentions with 5,284 distinct entities).

Our experimental evaluation appears in Table~\ref{tab:experiments}, which shows Precision, Recall and \Fonescore over the Bio-ID test set for different conditions. Experiments E1 and E2 show results of the NNER system trained over the full Bio-ID training dataset, which on the test set achieves \Fonescore of 82.99\% (BiLSTM+Softmax) and 83.34\% (BiLSTM+CRF). This simulates the performance over a large amount of labeled data and is our \textit{gold standard upper limit}. For the remaining experiments, we train an NNER system over a small dataset, a randomly selected 3\% subset of the Bio-ID training dataset, which mimics a low resource setting of 1,150 sentences with 1,258 protein/gene tags. We call the NNER model trained on 3\% of the data as NNER-3\%. We use the NNER-3\% model to predict labels for unknown tokens (noisily, since its accuracy is not perfect). Then, we apply different data augmentation techniques over the remaining 97\% of the Bio-ID training dataset, which simulates the accessibility of a large unlabeled corpus.  

\begin{table*}[htbp]
\begin{tabular}{@{}ccccccccccccc@{}}
\toprule
 &  & \multicolumn{4}{c}{\textbf{Training Data}} & \begin{tabular}[c]{@{}c@{}}\textbf{Output}\end{tabular} & \multicolumn{6}{c}{\textbf{Results}} \\ \midrule 
 &  & \begin{tabular}[c]{@{}c@{}}\small{True} \\ \small{Labels}\end{tabular} & \begin{tabular}[c]{@{}c@{}}\small{Ref-Set}\\ \small{Lookup}\end{tabular} & \begin{tabular}[c]{@{}c@{}}\small{Predicted}\\ \small{Labels}\end{tabular} & \begin{tabular}[c]{@{}c@{}}\small{Iterative}\\ \small{Refinement}\end{tabular} &  & \multicolumn{2}{c}{\small{P}} & \multicolumn{2}{c}{\small{R}} & \multicolumn{2}{c}{\small{F1}} \\ \cmidrule(l){3-13}
\multirow{4}{*}{\begin{tabular}[c]{@{}c@{}}\rotatebox{90}{\textbf{\small{True Labels}}}\end{tabular}} & E1 & 100\% & No & No & No & Softmax & \multicolumn{2}{c}{78.73} & \multicolumn{2}{c}{87.73} & \multicolumn{2}{c}{82.99} \\
 & E2 & 100\% & No & No & No & CRF & \multicolumn{2}{c}{80.75} & \multicolumn{2}{c}{86.09} & \multicolumn{2}{c}{83.34} \\
 & E3 & 40\% & No & No & No & Softmax & \multicolumn{2}{c}{78.70} & \multicolumn{2}{c}{36.05} & \multicolumn{2}{c}{49.45} \\
 & E4 & 40\% & No & No & No & CRF & \multicolumn{2}{c}{68.51} & \multicolumn{2}{c}{51.31} & \multicolumn{2}{c}{58.67} \\ \midrule
\multicolumn{1}{c}{} & \multicolumn{1}{c}{} & \multicolumn{1}{c}{} & \multicolumn{1}{c}{} & \multicolumn{1}{c}{} & \multicolumn{1}{c}{} & \multicolumn{1}{c}{} & \multicolumn{3}{c}{\textbf{Seed}} & \multicolumn{3}{c}{\textbf{Seed + Augmentation}} \\ \midrule
 &  &  &  &  &  & \multicolumn{1}{c}{} & \multicolumn{1}{c}{\small{P}} & \multicolumn{1}{c}{\small{R}} & \multicolumn{1}{c}{\small{F1}} & \multicolumn{1}{c}{\small{P}} & \multicolumn{1}{c}{\small{R}} & \multicolumn{1}{c}{\small{F1}} \\ \cmidrule(l){8-13} 
\multirow{5}{*}{\begin{tabular}[c]{@{}c@{}}\rotatebox{90}{\textbf{\small{Mixed Labels}}}\end{tabular}} & E5 &  3\% + 40\%  & No & Yes (57\%) & Yes & Softmax & 67.60 & 79.14 & 72.91 & 68.37 & 89.66 & 77.58 \\
 & E6 & 3\% + 40\% & No & Yes (57\%) & Yes & CRF & 67.94 & 86.77 & 76.21 & 72.79 & 88.92 & 79.75 \\
 & E7 & 3\% & C1 & Yes (97\%)& Yes & Softmax & 69.71 & 75.96 & 72.70 & 61.60 & 84.71 & 71.33 \\
 & E8 & 3\% & C2 & Yes (97\%) & Yes & Softmax & 69.71 & 75.96 & 72.70 & {\bf 70.30} & {\bf 84.23} & {\bf 76.63}  \\
 & E9 & 3\% & C2 & Yes (97\%)& Yes & CRF & 69.71 & 75.96 & 72.70 & {\bf 71.03} & {\bf 85.74} & {\bf 77.70}  \\
\bottomrule
\end{tabular}
\caption{Experimental Evaluation.  {[}C1 = Exact search (P=59.23, R=18.66).  C2 = Removed words in English dictionary
and words less than 4 characters; case-insensitive search (P=90.20, R=39.35){]}}
\label{tab:experiments}
\end{table*}

Experiment E3 shows the results for a simple baseline where we train our NNER system over the 3\% seed combined with one true protein label per sentence for the remaining 97\% of the Bio-ID training dataset, which removes $\sim$60\% of the protein labels. This experiment simulates an augmentation method with perfect precision, but a recall of only 40\%. 
Experiment E4 uses a CRF in the model's output layer for the same scenario, which results in a $\sim$9 point increase on \Fonescore to reach $\sim$58\% (although precision suffers). 
Even in this somehow unrealistic scenario that includes many of the available labels, the overall performance is significantly diminished from the the system trained on 100\% of the data ($\sim$25 percentage points below in \Fonescore). 

Experiments E5 and~E6 show the effect of our iterative label refinement method. We first train NNER-3\% on the seed data. Then we combine the seed, with the perfect precision (but partial, 40\%) labels as in experiments E3 and E4, and with the noisy predicted labels for the remaining tokens on the training dataset. Surprisingly, training over only 3\% of the data already achieves a good \Fonescore of 72.91\% for the BiLSTM+Softmax architecture and 76.21\% for the BiLSTM+CRF architecture. When we retrain this base system iteratively on the remaining data, the accuracy of the predicted labels increases, which leads to an improvement of $\sim$3-4 percentage points in \Fonescore (to 77.58\% for the BiLSTM+Softmax and 79.75\% for the BiLSTM+CRF). Thus, the iterative label refinement method reduces the distance to the 100\% trained system (E1, E2) from 25 to 4 percentage points, which is a substantial improvement.

Table~\ref{tab:iterres} shows the evolution of the iterative label refinement procedure for experiments E5 and~E6. We train NNER-3\% (Iter 0) and use it to predict labels for unknown tokens repeatedly, which yields a jump in performance in the first iteration (Iter 1), since the predicted labels are informative, and then a more gradual improvement as the labels are increasingly refined.

  \begin{table}
        \centering
        \caption{Performance of iterative refinement (E5, E6).}
        \label{tab:iterres}
        \footnotesize{ 
        \begin{tabular}{@{}rrrrrrrr@{}}
        \toprule
        & \multicolumn{3}{c}{BiLSTM+Softmax} && \multicolumn{3}{c}{BiLSTM+CRF} \\
        \cmidrule{2-4} \cmidrule{6-8}
          Iter & P & R & F1 && P & R & F1\\ \midrule
            0 & 67.60 & 79.14 & 72.91 && 84.14 & 66.49 & 74.28\\
            1 & 68.47 & 85.61 & 76.08 && 67.94 & 86.77 & 76.21 \\
            2 & 68.92 & 86.54 & 76.73 && 68.73 & 88.59 & 77.41 \\
            3 & 68.86 & 87.78 & 77.18 && 68.69 & 88.91 & 77.51 \\
            4 & 69.13 & 88.11 & 77.47 && 70.26 & 88.18 & 78.21 \\
            5 & 69.13 & 88.00 & 77.43 && 69.48 & 88.78 & 77.95 \\
            6 & 68.91 & 88.59 & 77.52 && 70.09 & 88.79 & 78.34 \\
            7 & 68.44 & 88.38 & 77.15 && 70.35 & 89.63 & 78.83 \\
            8 & 68.26 & 89.29 & 77.37 && 69.73 & 89.41 & 78.36 \\
            9 & 68.01 & 89.02 & 77.11 && 69.30 & 89.88 & 78.26 \\
            10 & 68.37 & 89.66 & \textbf{77.58} && 72.29 & 88.92 & \textbf{79.75}\\
        \bottomrule
        \end{tabular}
        }
  \end{table}
  
Finally, the remaining experiments (E7, E8, E9) simulate the more realistic scenario we seek, where we search for sentences in a large corpus to be labeled automatically. 
In experiment E7, we simply use our reference set to directly search for exact mentions in the corpus. Specifically, we search in a case sensitive way for protein/gene names from UniProt in the 97\% dataset that simulates our large unlabeled corpus (condition C1 in Table~\ref{tab:experiments}). 
Matching tokens are tagged as protein mentions for training. Since we know the true labels, we can compute the precision (=59.23\%) and recall (=18.66\%) of this selection technique, which in fact, is quite poor. 
The iterative training technique, which produced good results in the previous experiments, shows a decreased  performance (\Fonescore=72.70 for NNER-3\% and 71.33\% when using the augmented dataset). 
The low precision of the reference set search produces low quality augmented data, which in turn reduces performance.

For experiments E8 and~E9, we refined the reference set search to increase its precision. After error analysis, we discovered that many protein names were ambiguous. For example, the token ANOVA is a name of protein Q9UNW in UniProt, and also a well-known statistical procedure. Thus, we removed all protein names that appear in an English dictionary from our search. More drastically, we also removed protein names with fewer than 4 characters to avoid capturing acronyms that may not really be protein names. 
Finally, we also relaxed the matching strategy to be case insensitive and also to allow for partial matches (condition C2 in Table~\ref{tab:experiments}). 
For example, when searching for TIGAR, we will accept "Flag-tagged-TIGAR".
This selection technique improves precision to 90.20\%) and recall to 39.35\% on identifying correct proteins in Bio-ID. 
We constructed our augmented training dataset combining the seed, the reference-set matches, and the labels predicted by NNER-3\%, and applied our iterative bootstrapping procedure. This method achieves a \Fonescore of 76.63\% for BiLSTM+Softmax and of 77.70\% for BiLSTM+CRF. 

In summary, through these experiments we show that using a small labeled dataset and our automatic data augmented bootstrapping procedure (experiments E8, E9), we achieve a performance approaching that of a system trained with over 30 times more labeled data.

%%%%%%%%%%%%%%%%%%%
%%% DISCUSSIONS %%%
%%%%%%%%%%%%%%%%%%%
\section{Conclusion and Future Directions}

We proposed a method to improve NER with limited labeled data, which is often the case in technical domains, such as biomedicine. Our method combines bootstrapping and weakly-labeled data augmentation by using a small fully-labeled seed dataset and a large unlabeled corpus, automated labelling using a reference set, and an iterative label refinement process. Our experimental evaluation shows performance equivalent to systems trained with an order of magnitude more labeled data.

In future work, we aim to explore additional bootstrapping methods for other challenging datasets. 
We plan to apply the findings of these controlled experiments to a much larger, in-the-wild scenario where we use all the available labeled data as the seed and operate over a large corpus (e.g., all of PubMed, PubMed Central) to improve  state-of-the-art biomedical NER performance.

\balance

\bibliography{iclr}
\bibliographystyle{icml2019}

\end{document}